\documentclass{ldr-article}
\usepackage{biblatex}

\usepackage{tikz}
\usetikzlibrary{shapes, arrows}
\usepackage[obeyFinal]{easy-todo}
\usepackage{orcidlink}

\title{Computer-Aided Cytology Diagnosis in Animals: CNN-Based Image Quality Assessment for Accurate Disease Classification}

\author{
Jan Krupiński\textsuperscript{1}, Maciej Wielgosz\textsuperscript{1}\orcidlink{0000-0002-4401-2957}, Szymon Mazurek\textsuperscript{1,3}\orcidlink{0009-0006-7557-0157}, Krystian Strza\l{}ka\textsuperscript{1} \\
Pawe\l{} Russek\textsuperscript{1}, Jakub Caputa\textsuperscript{1}\orcidlink{0000-0002-9097-8269}, Daria \L{}ukasik\textsuperscript{1}, Jakub Grzeszczyk\textsuperscript{1} \\
Micha\l{} Karwatowski\textsuperscript{1}\orcidlink{0000-0001-6285-136X}, Rafa\l{} Fraczek\textsuperscript{1}\orcidlink{0000-0002-4311-2083}, Ernest Jamro\textsuperscript{1}\orcidlink{0000-0003-4632-2470}, Marcin Pietro\'n\textsuperscript{1,2}\orcidlink{0000-0001-9357-9231} \\
Sebastian Koryciak\textsuperscript{1,2}\orcidlink{0000-0001-6810-6897}, Agnieszka Dąbrowska-Boruch\textsuperscript{1}\orcidlink{0000-0003-2900-5668}, Kazimierz Wiatr\textsuperscript{1,2} \\\\
\textsuperscript{1} ACC Cyfronet AGH, Cracow, Poland \\
\textsuperscript{2} AGH University of Science and Technology, Cracow, Poland \\
\textsuperscript{3} Sano Centre for Computational Medicine, Cracow, Poland \\
}

\begin{document}

    \maketitle

    \begin{abstract}
      This paper presents a computer-aided cytology diagnosis system designed for animals, focusing on image quality assessment (IQA) using Convolutional Neural Networks (CNNs). The system's building blocks are tailored to seamlessly integrate IQA, ensuring reliable performance in disease classification. We extensively investigate the CNN's ability to handle various image variations and scenarios, analyzing the impact on detecting low-quality input data. Additionally, the network's capacity to differentiate valid cellular samples from those with artifacts is evaluated. Our study employs a ResNet18 network architecture and explores the effects of input sizes and cropping strategies on model performance. The research sheds light on the significance of CNN-based IQA in computer-aided cytology diagnosis for animals, enhancing the accuracy of disease classification.
    \end{abstract}
    
    \keywords{Computer-Aided Diagnosis; Cytology; Animals; Image Quality Assessment; Convolutional Neural Networks; Disease Classification; Veterinary Medicine.}
    \correspondingauthor{Jan Krupiński, j.krupinski@cyfronet.pl}

\section{Introduction}
Cytology \cite{simeonov2010fna} plays a vital role in veterinary medicine, enabling the early detection and diagnosis of various diseases through the examination of cellular samples. As digitialization becomes more prevalent in veterinary practice, the need for efficient and accurate computer-aided diagnostic systems has emerged. These systems have the potential to aid veterinarians in their assessments, improve diagnostic accuracy, and enhance the overall efficiency of disease classification in animals.

\begin{figure}
    \centering
    \includegraphics[scale=0.75]{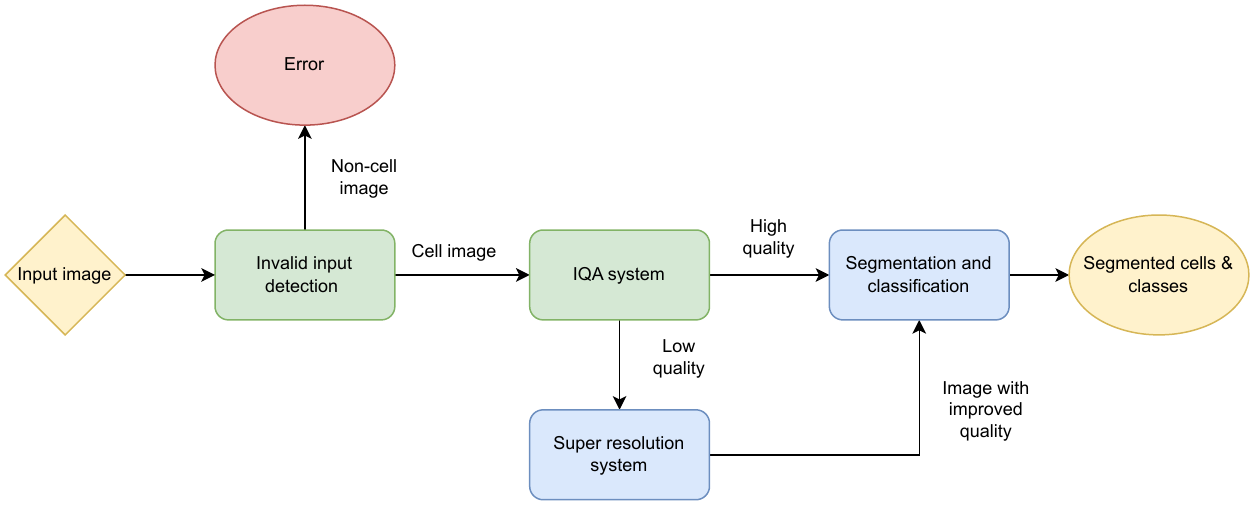}
    \caption{An overview of the discussed system for cytology images classification and segmentation, showing the end-to-end pipeline. Elements that are described in this paper are marked with green.}
    \label{fig:roadmap}
\end{figure}

The primary objective of this work is to contribute to the development of a computer-aided cytology diagnosis system with a specific emphasis on image quality assessment (IQA) using Convolutional Neural Networks (CNNs). Ensuring the quality of input images is critical for the optimal performance of deep learning models, particularly in the context of cytological image analysis, where subtle cellular features and abnormalities can significantly impact diagnostic accuracy. The system's diagram is shown in Figure \ref{fig:roadmap}.

This research presents three main contributions:

\begin{itemize}
\item We propose a set of building blocks tailored for a computer-aided cytology diagnosis system. These building blocks are designed to effectively incorporate image quality assessment into the system's pipeline, ensuring robust and reliable performance.
\item We comprehensively investigate the performance of CNNs in the task of image quality assessment on cytological images, analyzing various scenarios and image variations that can influence the model's ability to detect and handle low-quality input data accurately.
\item We assess the network's capability to differentiate between valid cellular samples and those containing artifacts or other invalid information, a critical step to ensure trustworthy diagnostic outcomes.
\end{itemize}

In the broader field of medical image analysis, image quality assessment has garnered significant attention. Numerous studies have demonstrated that CNNs can be sensitive to image distortions, such as blur and noise, which are common in cytological images of animals. Therefore, it is crucial to develop effective IQA methods for accurate cytology diagnosis in veterinary practice.

To achieve our goals, we adopt a simple ResNet18 \cite{he2015resnet} network architecture as the backbone of our computer-aided cytology diagnosis system. ResNet18 is selected for its reported ability to handle noisy images effectively, thanks to the incorporation of skip connections between convolutional layers. Additionally, we investigate the impact of varying input sizes and cropping strategies on the model's performance to determine the optimal configuration for our system.

The remainder of this paper is organized as follows: Section 2 provides an overview of related work in the field, emphasizing the significance of IQA in medical image analysis. Section 3 presents the datasets used for training and evaluation, focusing on cytological images of animals. In Section 4, we detail the network architecture and experimental setup. The results of our experiments are presented in Section 5, followed by a thorough discussion and conclusion in Sections 6 and 7, respectively.

\section{Related work}

Image Quality Assessment (IQA) is a topic of significant interest given its relevance to a multitude of fields, including deep learning. The quality of images plays a pivotal role in the performance of deep learning models. 

Research shows that Convolutional Neural Networks (CNNs) are notably sensitive to image distortions, particularly blur and noise \cite{dodge2016iqrdl}. Such distortions can significantly hinder the model's performance. This effect becomes even more pronounced in the medical field, where maintaining high standards of performance is crucial. 

In the context of melanoma classification it was noted that low-quality images result in reduced detection accuracy of malignant skin lesions \cite{bilge2021iqmelanoma}. Authors further highlighted the significant impact of blur and noise on the performance of ResNets \cite{he2015resnet}, a type of CNN that includes skip connections. Despite these image distortions, ResNets are found to be relatively resilient.

Interestingly, one work discovered that even minor alterations in input resolution could influence the performance of the network in classifying endoscopy images \cite{Thambawita2021imageendoscopy}. This echoes the findings of other research, where authors drew similar conclusions while examining the classification of X-ray images \cite{sabottke2020radiology}.

Deep learning methods have also found utility as IQA tools. Researchers utilized features learned by a CNN and an SVM to classify image quality \cite{bianco2016blindiqr}. A similar approach was employed in a work where authors supplemented CNN features with handcrafted features to predict the quality of eye fundus images, achieving an AUROC of 0.98 \cite{yu2017iqcdr}.

Comparable performance on the same kind of data was also achieved without the use of handcrafted features. This suggests that deep learning alone can be a powerful autonomous tool for IQA \cite{Tennakoon2016} . 

Finally, CNNs are capable of detecting artifacts in MRI scans and assessing the diagnostic quality of an image \cite{ma2020iqamedical}. This demonstrates that effective IQA with deep learning is feasible, even for complex imaging modalities.

\section{Materials and methods}
\subsection{Datasets}

\begin{figure}[h]
    \centering
    \includegraphics[scale=0.4]{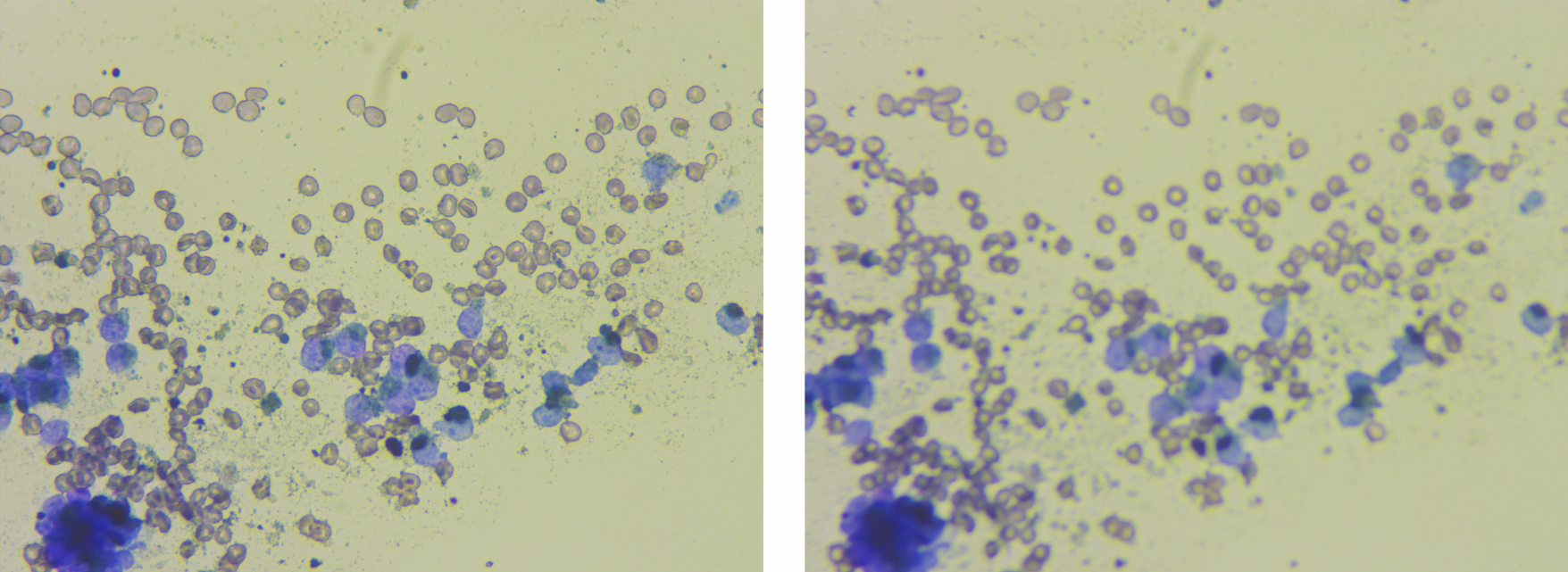}
    \caption{A sample from the first dataset: the high-quality image on the left, and the lower-quality one on the right.}
    \label{fig:ds-original}
\end{figure}

Two datasets were employed for this study. The first one comprised of 2,600 RGB canine cytological samples with a resolution of 2592x1944. Each sample was represented in two versions: a high-quality version, obtained with a properly focused microscope, and a lower-quality version, which was the result of a misaligned microscope focus. Examples of these can be observed in Figure \ref{fig:ds-original}.

The second dataset served the purpose of validating the input cell image. It was constructed by merging the cell images from the first dataset with an additional 21,234 RGB images of various sizes, sourced from the ImageNet dataset \cite{deng2009imagenet}. These additional images were chosen from classes that closely mimic real cell images, with further details provided in the Appendix.

In order to render the dataset more practical, we incorporated examples of cells with dark edges. Such images are frequently encountered in online systems due to factors such as the lack of a stable camera mount or taking photos with smartphones. These images depict cells encircled by a dark ring and might contain random, blurry objects at the edges. We generated these images by adding a dark surrounding circle to each photo from the first dataset.

\begin{figure}[h]
    \centering
    \includegraphics[scale=0.3]{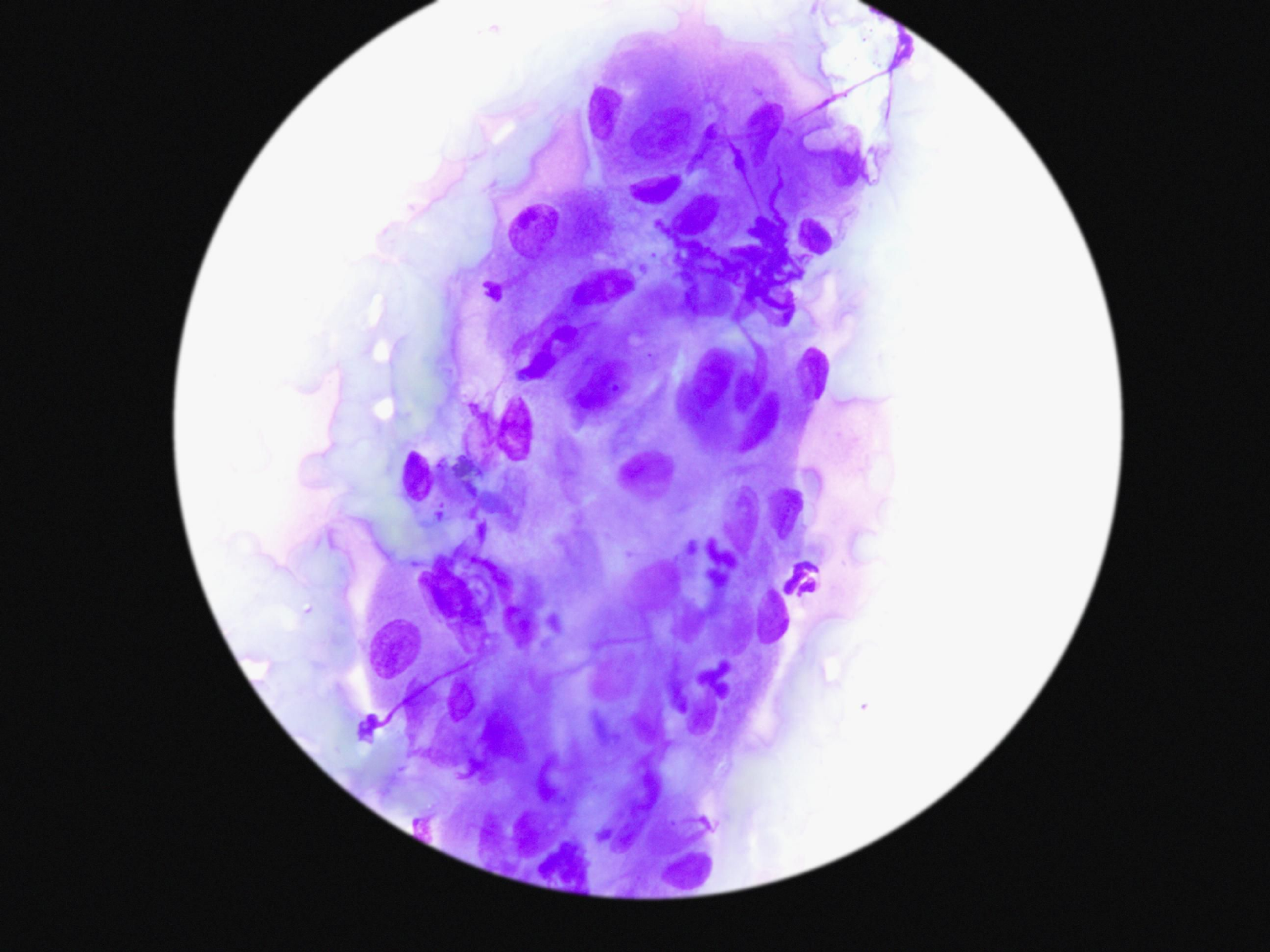}
    \caption{An example of a generated sample with dark edges.}
    \label{fig:dark-edges}
\end{figure}

Figure \ref{fig:dark-edges} showcases an example of the image with dark edges that were included in the dataset. Consequently, the updated dataset encompasses a total of 31,634 samples, which consist of 10,400 cell samples and 21,235 non-cell samples.

\subsection{Network architecture}
For the purpose of this study, simple ResNet18 \cite{he2015resnet} network was chosen. The choice was based on the reported ability of the network to handle noisy images \cite{bilge2021iqmelanoma} due to skip connections between convolutional layers.
    
    

\subsection{Experimental setup}

In the subsequent experiments, we utilized a transfer learning approach to train the selected network. The initial parameters were derived from training the network on the ImageNet dataset. We assessed the performance using several metrics, including accuracy, F1-score, precision, and recall. The binary cross-entropy loss function was optimized using Stochastic Gradient Descent (SGD) with momentum \cite{sutskever2013momentum}. A learning rate of $10^{-4}$ and momentum of 0.9 were set for this process.

Unless stated otherwise, all input images were resized to 224x224 using bilinear interpolation. We opted for a batch size of 16 samples. The experiments were evaluated using a 5-fold cross-validation strategy, where the mean performance across all folds was calculated for each trained model. 

An additional 15\% of the data was set aside from the training subset to form the validation subset. This subset was used in an early stopping protocol: if no improvement in validation loss was observed for a predetermined number of epochs (10), the training was terminated. Test inference was carried out using the weights from the epoch that exhibited the lowest validation loss.

\subsection{The Toolset: Software Frameworks and Hardware Utilized}
Our experiments were conducted utilizing Python 3.9.14 and leveraged the functionalities of PyTorch 2.0.1 \cite{paszke2019torch}. The model, along with its corresponding weights, were sourced from the repository of the torchvision 0.15.2 library.

To handle the computational demands, we employed the power of the Athena Supercomputer Cluster, utilizing six AMD EPYC 7742 64-Core CPUs and one Nvidia A-100 GPU for this purpose. The GPU operations were facilitated by the CUDA 11.7 drivers.

\section{Results}

\subsection{Establishing baseline performance and the problem of splitting the data and shuffling}

First, we explored whether the pairs of high-quality and low-quality images could be split when creating training, testing, and validation subsets. Two strategies were compared: the first strategy ensured that both the high-quality and low-quality images from the same sample were included in a given subset (Sameidx), while the second strategy avoided including images from the same sample in the same subset (Diffidx). The shuffling was also taken into account for Sameidx approach - we evaluated if maintaining the pairs within a single batch also influences the results. Outcomes are presented in Table \ref{tab:init-idx}.



\begin{table}[h]
\centering
\begin{tabular}{p{2.2cm}|p{1.8cm}|p{1.8cm}|p{1.8cm}|p{1.4cm}|p{1.1cm}|p{1.1cm}}

Experiment          & Accuracy [\%] & F1 Score [\%] & Precision [\%] & Recall [\%] & Total Time [min] & Avg. Time [min]  \\ \hline
Diffidx, No Shuffle & 96.21 \newline $\pm$ 0.93 & 96.17 \newline $\pm$ 0.95 & 97.1 \newline $\pm$ 1.5 & 95.3 \newline $\pm$ 1.8 & 380 & 38 \newline $\pm$ 32 \\ \hline
Sameidx, No Shuffle & 96.08 \newline $\pm$ 0.51 & 96.03 \newline $\pm$ 0.49 & 97.1 \newline $\pm$ 1.0 & 94.96 \newline $\pm$ 0.50 & 100 & 20.0 \newline $\pm$ 1.9  \\ \hline
Sameidx, Normal Shuffle & 97.19 \newline $\pm$ 0.45 & 97.17 \newline $\pm$ 0.99 & 98.13 \newline $\pm$ 0.98 & 96.23 \newline $\pm$ 0.32 & 184 & 36.8 \newline $\pm$ 4.2  \\ \hline
Sameidx, Pair Shuffle   & 97.08 \newline $\pm$ 0.32 & 97.04 \newline $\pm$ 0.33 & 98.16 \newline $\pm$ 0.62 & 95.96 \newline $\pm$ 0.76 & 136 & 27.2 \newline $\pm$ 3.9  \\ 
\end{tabular}
\caption{Results evaluating various splitting and shuffling strategies.}
\label{tab:init-idx}
\end{table}

Interestingly, keeping the pairs together in the subsets resulted in marginally less accuracy but showed more stability in performance across the folds. Notably, the average training time per fold was nearly halved compared to strategies that did not maintain the paired samples. Similar observations were made when assessing shuffling methods — the strategy that preserved pairs within the batches led to substantially quicker training with equivalent performance. For subsequent experiments, we utilized the Sameidx, Pair Shuffle strategy unless otherwise specified.

\subsection{Evaluating the effect of input size}
We suspected that input size would have an impact on the model's performance, therefore the experiment evaluating training on images with the original size was conducted. We decided to leverage the findings from previous experiment, using paired data split to shorten the training process. The results are shown in Tab. \ref{tab:no-resize} along with the same experiment with smaller input size from previous runs for comparison. 
\begin{table}[h]
\centering
\begin{tabular}{p{1.8cm}|p{2.4cm}|p{2.2cm}|p{2.5cm}|p{2.2cm}|p{1.9cm}}

Input Size & Accuracy [\%] & F1 Score [\%] & Precision [\%] & Recall [\%] & Avg. Time [min]  \\ \hline

244x244   & 97.08 $\pm$ 0.32 & 97.04 $\pm$ 0.33 & 98.16 $\pm$ 0.62 & 95.96 $\pm$ 0.76 & 27.2 $\pm$ 3.9  \\ 
2592x1944 & 98.54 $\pm$ 0.22 & 98.53 $\pm$ 0.23 & 98.80 $\pm$ 0.34 & 98.27 $\pm$ 0.65 & 164 $\pm$ 53  \\
\end{tabular}
\caption{Comparison of the results using original and decreased input size.}
\label{tab:no-resize}
\end{table}

It can be seen that using the original input size increases the performance on average, however it also increased the total run time by more than 8 times. Therefore, we decided that performance gain is not worth the increased computational costs, and decided to proceed with further experiments using 244x244 input size. 
\subsection{Random cropping and it's influence on model's performance}
\begin{figure}[bt]
    \centering
    \includegraphics[scale=0.8]{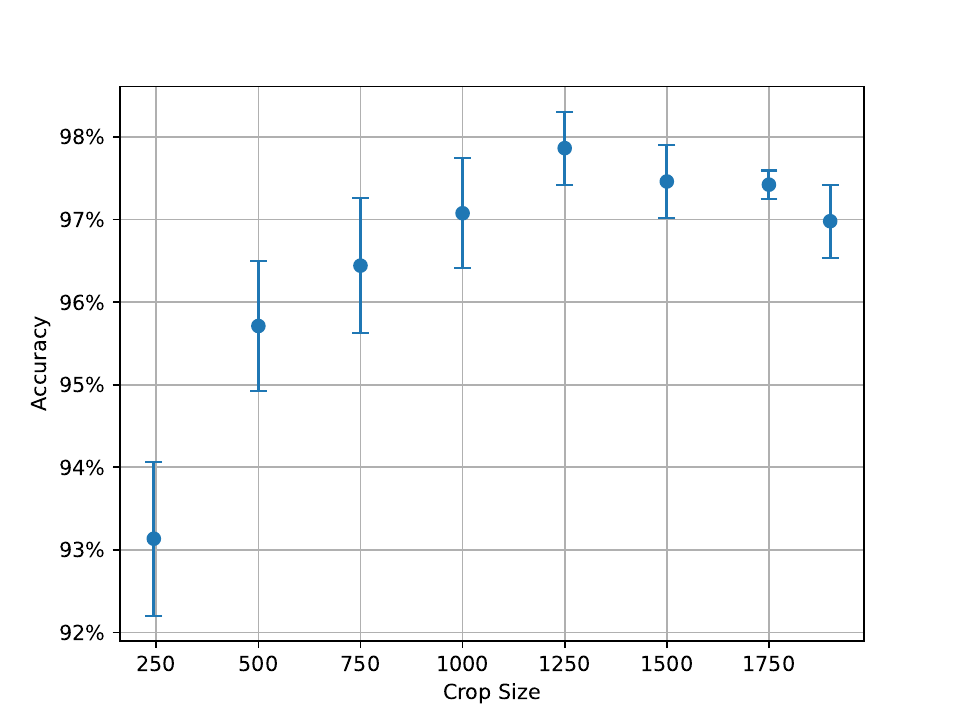}
    \caption{Average accuracy of the model across folds using different crop sizes.}
    \label{fig:crop-acc}
\end{figure}

\begin{figure}[bt]
    \centering
    \includegraphics[scale=0.8]{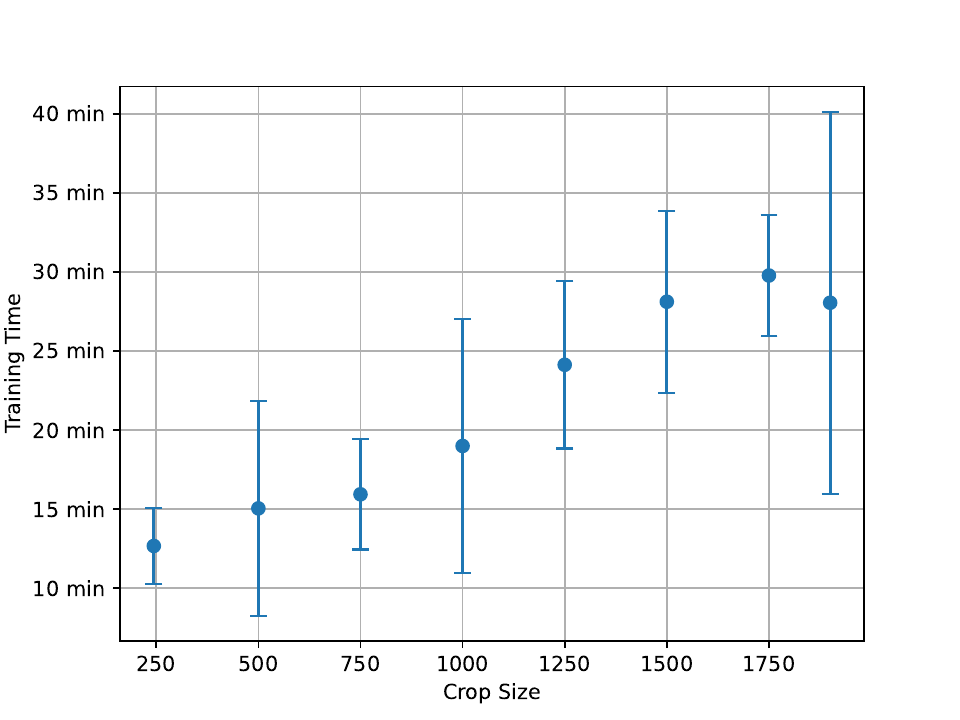}
    \caption{Time to train models for all folds using different crop sizes.}
    \label{fig:crop-time}
\end{figure}

When performing the random crop, there exists a chance that the cropped area will not contain enough informative features (i.e. only background of the image will be visible without cells present). Also, crop undergoes smaller compression when resizing to the lower input resolution, therefore some relevant features can remain unaffected. Therefore, in the next experiment the network was trained and tested on randomly cropped images with varying crop sizes. The results are shown in Figure \ref{fig:crop-acc} with corresponding average training time in Figure \ref{fig:crop-time}.

The best results were achieved for 1250x1250 random crop, with accuracy of $97.87\pm 0.44$ \%  and F1-score of $97.85\pm 0.44$ \%  and average training time of $1447 \pm317$ seconds.

\subsection{Sensitivity to the magnification}
\begin{figure}[bt]
    \centering
    \includegraphics[scale=0.8]{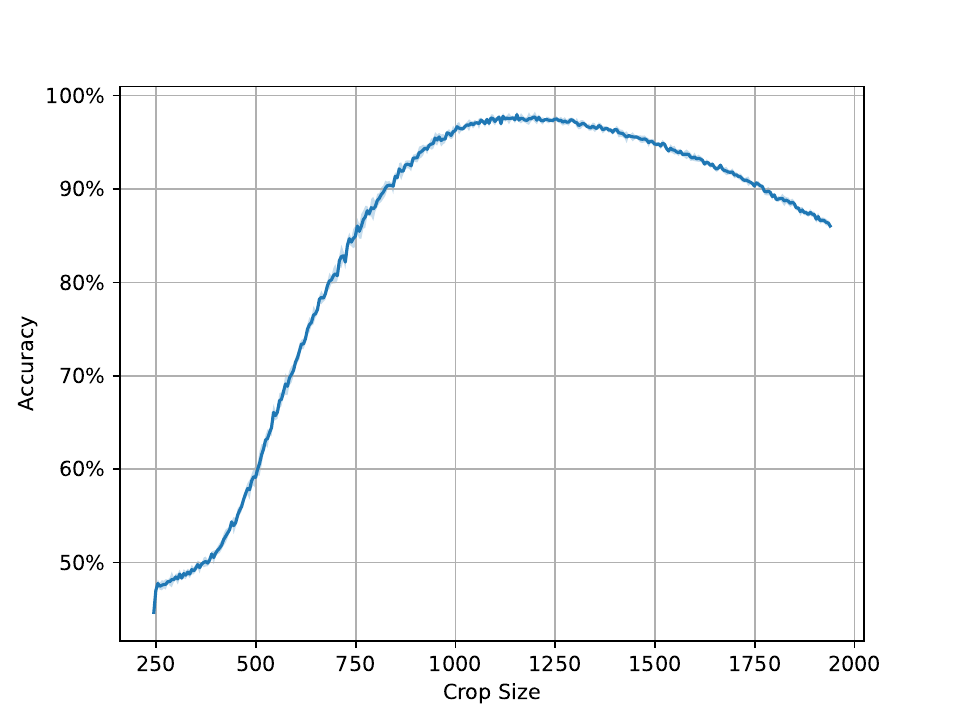}
    \caption{The performance of the model trained on 1250x1250 random crops when tested on samples cropped with different crop sizes.}
    \label{fig:mag-sens}
\end{figure}

Sensitivity to magnification is relevant due to the fact that images evaluated by a widely accessible system usually are taken using different equipment, which possibly can affect the system's performance. To assess this, we trained the model on samples obtained by 1250x1250 crops of original images, using 80\% images for training and 20\% for testing.  The training dataset was further split and 15\% of the subset was subtracted for validation. The crop size was chosen as the best compromise between accuracy and time efficiency. The trained model was tested on the holdout set that was cropped with different crop sizes. The results are presented in Figure \ref{fig:mag-sens}.

It is clearly visible that indeed model performs the best when tested on the random crop size that is close to the one that was used for training. The performance slowly decreases when the crop size is increases and drops sharply when the crop size is lowered from the training value.

\subsection{Assessing the quality of the crop}
\begin{figure}[bt]
    \centering
    \includegraphics[scale=0.1]{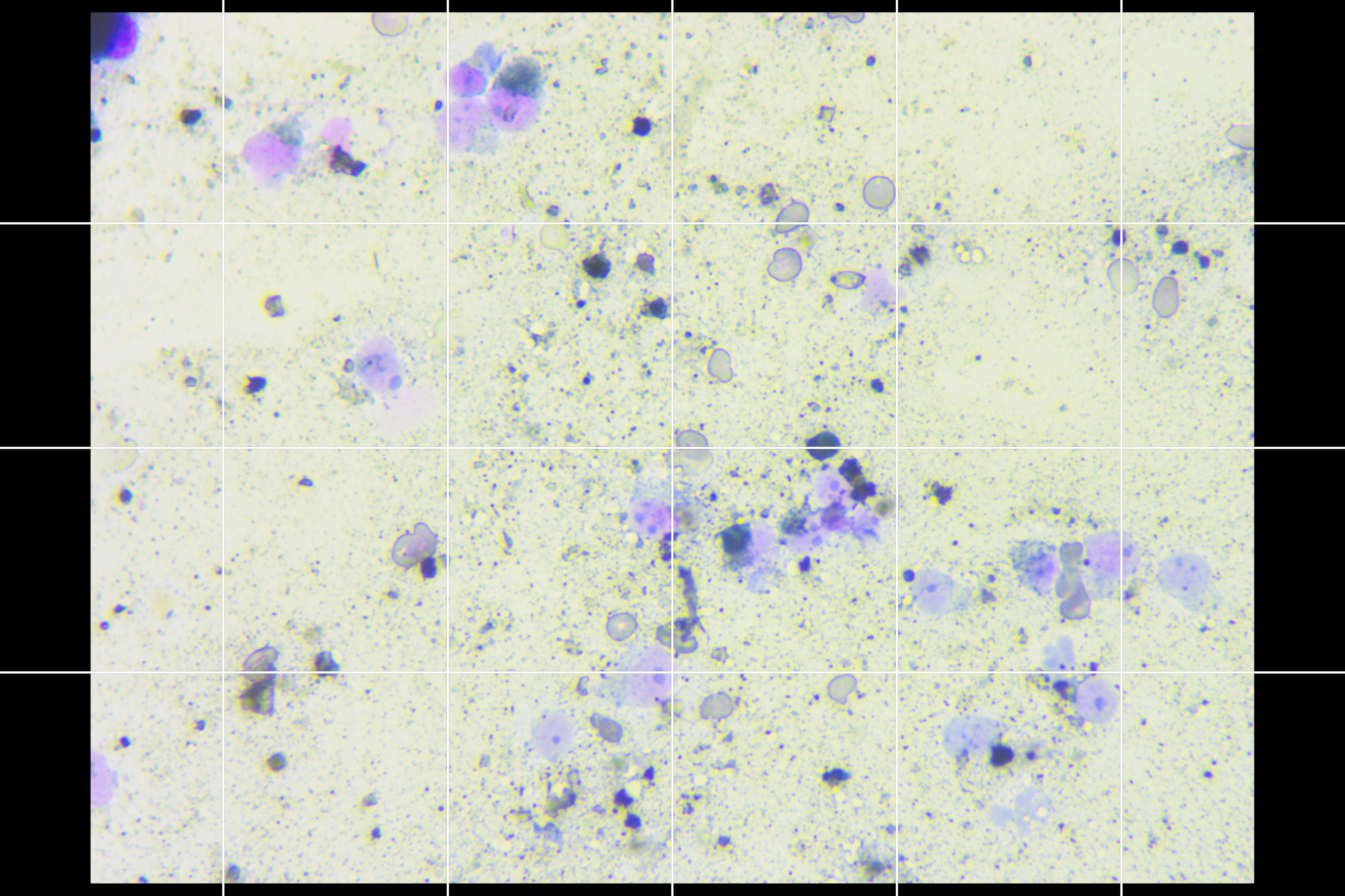}
    \caption{An example of image divided into patches. The presented case shows the version taking all fragments into account, padding the edge fragments with zeroes.}
    \label{fig:slices-ensemble}
\end{figure}

The crops are done randomly, therefore there is a risk that a cropped part of the image will not contain enough details for the network to make a correct prediction. For this task, a model trained during the previous experiment on 500x500 crops was used. The remaining test samples were cropped into equal 500x500 patches, which were then passed through the network to obtain prediction for every patch. We chose smaller crop size than the optimal due to the fact of increased number of patches extracted per image. Example of sliced image is shown in Figure \ref{fig:slices-ensemble}. The obtained probabilities from prediction on every patch were combined, using different methods of calculating the final prediction score. The following approaches were tested:
\begin{itemize}
    \item \textbf{Control} - single random 500x500 crop
    \item \textbf{Sum} - image was divided into equal 500x500 fragments, all edge fragments which were smaller were not taken into account. The outputs were summed.
    \item \textbf{Sum with size weights} (Sum \& sw.)  - image was divided into 500x500 fragments, with the smaller edge fragments being padded with zeroes. The outputs were summed, with a weight corresponding to the percent of the fragment that wasn't padding.
    \item \textbf{RGB variance} (RGB var.) - image was divided into equal 500x500 fragments, all edge fragments which were smaller were not taken into account. The outputs were summed, with the weight being the average variance of the RGB channels of the image fragment.
    \item \textbf{RGB variance with size weights} (RGB var. \& sw.) - image was divided into 500x500 fragments, with the smaller edge fragments being padded with zeroes. The outputs were summed, with a weight being the average variance of the RGB channels of the image fragment multiplied by the percent of the fragment that wasn't padding.
    \item \textbf{Saturation variance} (Saturation var.) - image was divided into equal 500x500 fragments, all edge fragments which were smaller were not taken into account. The outputs were summed, with the weight being the variance of the saturation of the image fragment.
    \item \textbf{Saturation variance with size weights} (Saturation var \& sw.)- image was divided into 500x500 fragments, with the smaller edge fragments being padded with zeroes. The outputs were summed, with a weight being the saturation variance of the image fragment multiplied by the percent of the fragment that wasn't padding. 
\end{itemize}
The results are shown in Tab.\ref{tab:ensemble-results}. Used methods have increased the model's performance on the test set with the \textit{RGB variance} method performing the best and the ones not taking edge fragments into account being nearly equally good. The time to perform the inferences was longer for the methods based on ensemble prediction, which is expected as more photos need to be evaluated. 

\begin{table}[h]
\centering
\begin{tabular}{p{2.2cm}|p{1.6cm}|p{1.6cm}|p{1.7cm}|p{1.4cm}|p{1.5cm}|p{1.6cm}}
Experiment  &  Accuracy [\%]   & F1 Score [\%]  & Precision [\%] & Recall [\%] & Total Time [s] & Avg. Time [s]\\ \hline
Control & 95.71 \newline $\pm$ 0.79 & 95.73 \newline $\pm$ 0.78 & 95.35 \newline $\pm$ 0.99 & 96.12 \newline $\pm$ 0.99 & 275 &  54.8 \newline $\pm$ 3.1\\ \hline
Sum & 98.10 \newline $\pm$ 0.42 & 98.09 \newline $\pm$ 0.41 & 98.23 \newline $\pm$ 0.89 & 97.96 \newline $\pm$ 0.60 & 556 & 111.2 \newline $\pm$ 1.6\\ \hline
Sum \& sw. & 97.98 \newline $\pm$ 0.62 & 97.97 \newline $\pm$ 0.20 & 98.60 \newline $\pm$ 0.65 & 97.34 \newline $\pm$ 0.52 & 654 & 130.8 \newline $\pm$ 3.2\\ \hline
RGB var & 98.44 \newline $\pm$ 0.26 & 98.43 \newline $\pm$ 0.26 & 98.95 \newline $\pm$ 0.53 & 97.92 \newline $\pm$ 0.63 & 519 & 103.9 \newline $\pm$ 2.3\\ \hline
RGB var \&  sw. & 97.69 \newline $\pm$ 0.25 & 97.67 \newline $\pm$ 0.27 & 98.41 \newline $\pm$ 0.78 & 96.96 \newline $\pm$ 1.14 & 678 & 135.5 \newline $\pm$ 4.0\\ \hline
Saturation var  & 98.23 \newline $\pm$ 0.22 & 98.22 \newline $\pm$ 0.22 & 98.76 \newline $\pm$ 0.70 & 97.69 \newline $\pm$ 0.61 & 541 & 108.2 \newline $\pm$ 5.2\\ \hline
Saturation var \&  sw. & 97.19 \newline $\pm$ 0.44 & 97.15 \newline $\pm$ 0.48 & 98.36 \newline $\pm$ 0.88 & 96.00 \newline $\pm$ 1.58 & 679 & 135.9 \newline $\pm$ 1.9\\ 
\end{tabular}
\caption{Comparison of the results using different methods of patch weighting in an ensemble prediction scenario.}
\label{tab:ensemble-results}
\end{table}

\subsection{Detection of invalid input objects}

In the last experiment, we evaluated the ability of the network to determine if a given input image is indeed a cell or not. For this purpose, the network was trained on the second dataset. As the dataset was imbalanced, the loss was modified to weigh samples of given classes with 0.67 and 0.33 for cell and non-cell images respectively. The model had reached excellent average accuracy of $99.99 \pm21$\%, F1-score of $99.99 \pm 15$\% and needed an average of $3408 \pm1788$ seconds to converge.

\section{Conclusions and discussion}

The conducted experiments have substantiated that deep learning vision models can be effectively employed for Image Quality Assessment (IQA) and input validity in computer-aided diagnosis systems. Utilizing Convolutional Neural Networks (CNNs) and transfer learning for image quality assessment tasks yields satisfying performance. This holds true even when the network is not state-of-the-art and its architecture and hyperparameters are not meticulously tuned for the task.

Interestingly, when dealing with a paired IQA task as in this study, preventing samples from being assigned to different data subsets during training can expedite convergence. Reducing the resolution of input images from their original size leads to only a slight decrease in performance, while significantly reducing training time. This insight could be particularly beneficial when training and deploying systems in environments with limited computational resources.

Implementation of random cropping can help to improve the results, increasing the model's accuracy. It can also be expanded by combining the inference results of many crops from the image to mitigate the risk of cropping parts of the image that is less relevant when making a prediction. This approach increases inference time, but only from 52.7 $\pm$ 3.0 \textit{ms} to 99.9 $\pm$ 2.2 \textit{ms} per image. This is an acceptable trade-off for medical systems such as the presented one, where each mistake can lead to serious consequences for the examined patient.

Furthermore, CNN also demonstrated efficacy in determining whether an image is a photo of a cell. This is especially crucial in open systems where users have the freedom to provide arbitrary inputs, potentially leading to unexpected model behavior and predictions.

Future extensions of this study could include additional experiments to identify the optimal architecture and hyperparameters, potentially improving system performance. Determining the optimal input resolution could also be worth exploring, as this might enhance results without increasing computational costs. 
Finally, an investigation into other methods of combining predictions from multiple crops of the image could lead to greater accuracy.
These could be valuable contributions to future solutions in this field.

\section*{Acknowledgements}
The work is partially supported by PRACE-LAB 2 project at ACK-Cyfronet AGH. Szymon Mazurek is partially supported by the European Union’s Horizon 2020 research and innovation programme under grant agreement Sano No 857533. Szymon Mazurek is partially supported by Sano project carried out within the International Research Agendas programme of the Foundation for Polish Science, co-financed by the European Union under the European Regional Development Fund.

This research was supported in part by the PLGrid infrastructure grant plglaoisi23 on the Athena supercomputer cluster.

\section*{Data availability}
The code used for experiments will be made available publicly on \url{https://git.plgrid.pl/projects/CYFROVET/repos/sr_classifier/browse}
\printbibliography
\section*{Appendix}

\subsubsection*{Choosing objects similiar to the cell images from ImageNet dataset}

As ImageNet is an enormous dataset, manual choice of images that can potentially be considered by the model as cells was impossible. We therefore decided for the following approach: the ResNet18 model pre-trained on the ImageNet-1K dataset was used to perform the inference on the canine cytological samples. Then, for every class, the output logits of the model were averaged across samples. We therefore obtained scorings for entire classes and chose 10 that on average had the highest logits values. They can be seen in Table \ref{tab:imagenet-samples}.

\begin{table}[h]
\centering
\begin{tabular}{l|l}

Label                               & Average logit              \\
\hline
jellyfish                           & 0.090 \\
\hline
cauliflower                         & 0.048 \\
\hline
bubble                              & 0.040 \\
\hline
Petri dish                          & 0.038 \\
\hline
nematode                            & 0.034 \\
\hline
velvet                              & 0.033 \\
\hline
jigsaw puzzle                       & 0.033 \\
\hline
ant                                 & 0.027 \\
\hline
handkerchief                        & 0.022 \\
\hline
poncho                              & 0.020 \\

\end{tabular}
\caption{Classes with highest average logit when tested using ResNet18 pre-trained on ImageNet-1K.}
\label{tab:imagenet-samples}
\end{table}

\end{document}